\documentclass[runningheads]{llncs}
\usepackage{graphicx}
\usepackage{tabularx}
\usepackage{multirow}
\usepackage{booktabs}
\usepackage{subcaption}

\usepackage{tikz}
\usepackage{comment}
\usepackage{overpic}
\usepackage{algorithm}
\usepackage{algpseudocode}
\usepackage{amsmath,amssymb} 
\usepackage{color}

\usepackage[accsupp]{axessibility}  

\usepackage{hyperref}
\hypersetup{
    colorlinks=true,
    linkcolor=blue,
    filecolor=blue,      
    urlcolor=blue,
    citecolor=cyan,
}

\begin{document}
\pagestyle{headings}
\mainmatter
\def\ECCVSubNumber{100}  

\title{SketchSampler: Sketch-based 3D Reconstruction \\ via View-dependent Depth Sampling} 

\titlerunning{SketchSampler}
%
\author{Chenjian Gao\inst{1} \and
Qian Yu$^{1\!}$\thanks{Corresponding author.} \and
Lu Sheng\inst{1} \and
Yi-Zhe Song\inst{2} \and
Dong Xu\inst{3}}
\authorrunning{Gao et al.}
%
\institute{School of Software, Beihang University\\
\email{\{gaochenjian, qianyu, lsheng\}@buaa.edu.cn}\\\and
SketchX, CVSSP, University of Surrey\\
\email{y.song@surrey.ac.uk}\\\and
Department of Computer Science, The University of Hong Kong\\
\email{dongxudongxu@gmail.com}}

\maketitle
\begin{abstract}

Reconstructing a 3D shape based on a single sketch image is challenging due to the large domain gap between a sparse, irregular sketch and a regular, dense 3D shape. Existing works try to employ the global feature extracted from sketch to directly predict the 3D coordinates, but they usually suffer from losing fine details that are not faithful to the input sketch.
Through analyzing the 3D-to-2D projection process, we notice that the density map that characterizes the distribution of 2D point clouds(i.e., the probability of points projected at each location of the projection plane) can be used as a proxy to facilitate the reconstruction process. 
To this end, we first translate a sketch via an image translation network to a more informative 2D representation that can be used to generate a density map. Next, a 3D point cloud is reconstructed via a two-stage probabilistic sampling process: first recovering the 2D points(i.e., the $x$ and $y$ coordinates) by sampling the density map; and then predicting the depth(i.e., the $z$ coordinate) by sampling the depth values at the ray determined by each 2D point.
Extensive experiments are conducted, and both quantitative and qualitative results show that our proposed approach significantly outperforms other baseline methods. Code has been released: \url{https://github.com/cjeen/sketchsampler}

\end{abstract}

\section{Introduction}

 \begin{figure}[t]
  \centering
  \begin{overpic}[width=0.7\linewidth]{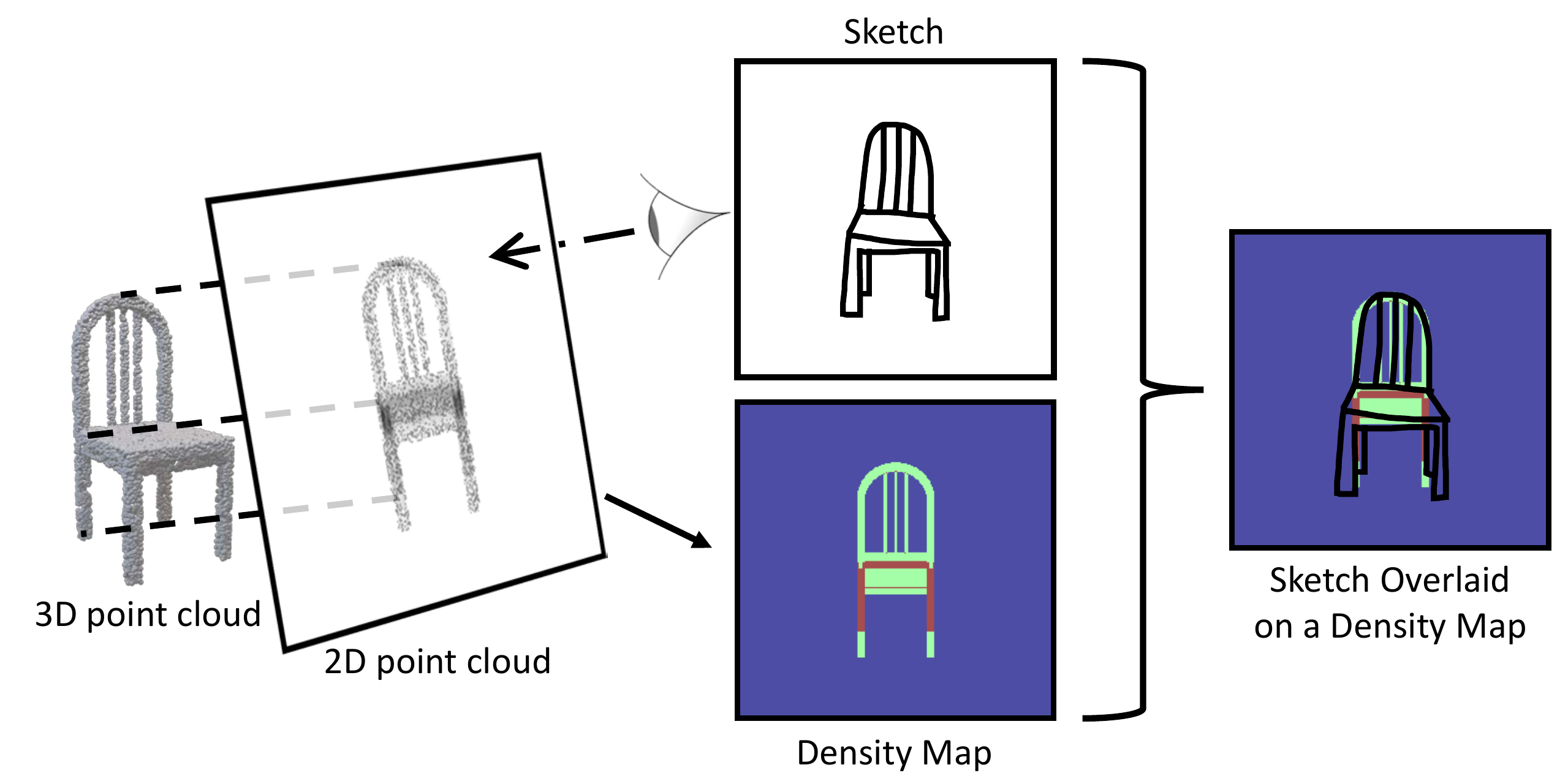}
  \end{overpic}
  \caption{The motivation of our work. We can see: 1) a 2D point cloud can be generated by projecting a 3D shape onto an image plane. On the projection plane, some locations have more than one 2D point with different depth values. 2) The distribution of 2D points can be characterized by a density map, where the value at each location indicates the probability of points projected at that location. `Red' color indicates higher density. 3) The density map is spatially rough-aligned with the sketch.}
\label{fig:motivation}
\end{figure}
Sketching is an intuitive approach for humans to express their ideas, and it has been adopted for 3D modeling for decades. With the rapid development of deep learning and virtual reality (VR) techniques, sketch-based 3D modeling has attracted increasing attention from both academia and industry~\cite{li2017bendsketch,Xu:2014:True2Form,li2018robust,delanoy20183d,lun20173d}, showing great potential in designing, animation, and entertainment. 

In recent years we have witnessed great progress in sketch-based 3D modeling. Motivated by the success of image-based single-view 3D reconstruction (SVR), most sketch-based 3D modeling approaches follow a well-known pipeline of SVR \cite{DBLP:journals/corr/FanSG16,wang20203d}, which firstly encodes a sketch into a feature vector with a convolution neural network (CNN), and then utilizes multilayer perception (MLP) based decoders to generate a fixed number of 3D coordinates that define the point cloud of a 3D shape. 
Considering that sketch is usually sparse and ambiguous, global feature is a reasonable sketch representation \cite{zhang2021sketch2model}, as it summarizes the sketch in a coarse level(e.g., semantic category and its conceptual shape). However, it is hard for a model to reconstruct a 3D shape with fine details from a global feature due to the huge domain gap between a sketch and a 3D shape. 

Fig.~\ref{fig:motivation} illustrates a 2D point cloud projected from a 3D shape. Note that on the projection plane, there could exist more than one 2D points with different depth values at the same location because of occlusions. The distribution of a 2D point cloud can be characterized by a density map, indicating the probability of points projected at each location. In other words, if we have a density map, we can infer the corresponding 2D point cloud. 
It is interesting to see that a sketch is spatially rough-aligned with the density map. Considering that both sketch and density map are 2D images, their domain gap is supposed to be smaller than that between sketch and 3D shape. This motivates us to introduce the density map as the proxy to facilitate sketch-to-3D reconstruction. Namely, given an input sketch, the reconstruction model first generates a density map to recover a 2D point cloud (i.e., $x$ and $y$ coordinates) and then predicts the depth value for each 2D point (i.e., $z$ coordinate). \textit{From a probabilistic view}, this reconstruction process can be interpreted as predicting the joint distribution of $x,y,z$ coordinates from a 3D shape, which defines a two-stage sampling process. Specifically it \textit{firstly} samples $x,y$ coordinates from the distribution of $P(X,Y|I)$ generated from the sketch $I$, and \textit{secondly} samples $z$ coordinate for each $(x,y)$ location from the distribution of $P(Z|x,y,I)$. 
Note that the distribution of $P(X,Y|I)$ is a 2D point cloud to be generated from a sketch.

However, considering the sparsity and ambiguity characteristics of sketch, there is also a domain gap between a sketch and a density map. As displayed in Fig.~\ref{fig:motivation}, the visible and occluded object surfaces, and the vacancy between surfaces can all be shown as blank in a sketch. We thus adopt an image translation model~\cite{isola2017image} to complete the missing information while preserving the spatial information in a sketch before predicting the density map. 

In this work, we present a new method for sketch-based SVR, which consists of two components: a sketch translator and a point cloud generator. The sketch translator adopts a CNN-based encoder-decoder network, where the encoder network extracts features from the input sketch and the decoder network infers target 3D information and outputs a more informative 2D representation. 
Based on the output of the sketch translator, our point cloud generator aims to reconstruct a point cloud of the corresponding 3D shape. It first predicts the density map which can be used as guidance to recover 2D point clouds, 
and then samples along a ray determined by each 2D projected point to predict depth values, where the point with farther depth values means it is occluded by the point with nearer depth values.

It is worth noting that a sketch may exhibit different levels of deformation and abstraction. Here we focus on sketches with reliable shape and fine-grained details, i.e., sketches with significant deformation or only expressing conceptual ideas are not considered in this work.
To demonstrate the effectiveness of our proposed model, we train and test it on a newly rendered dataset, \textit{Synthetic-LineDrawing}. 
The contributions of this work are three-fold:
\begin{itemize}
    \item First, we present a novel method for sketch-based single-view 3D reconstruction, in which a 3D shape is recovered in two easier but indispensable steps, sketch translation and point cloud generation;
    \item Second, we formulate the point generation process as a two-stage probabilistic sampling process, where the density map is introduced as a guidance.
    Besides, an image translation model is used for sketch translation to preserve the spatial information in a sketch and to further reduce domain gap;
    \item Third, the proposed model can reconstruct a 3D shape faithful to the sketched object. Its effectiveness has been demonstrated through extensive experiments on both synthetic and hand-drawn sketch datasets.
\end{itemize}

\section{Related Works}

\subsection{Deep Sketch-based 3D Modeling}
Sketch-based modeling is a problem that has been studied for a long time. The earlier methods predicted local geometric properties from hand-crafted rules and then inferred the 3D shape from the geometric properties~\cite{zhong2020towards,zhong2020deep}. In recent years, some deep learning based methods have been proposed for sketch-based 3D modeling. Wang \textit{et al} introduced a method to reconstruct 3D shapes based on retrieval~\cite{wang2018unsupervised}. The work~\cite{wang20203d} proposed to generate point clouds from a single hand-drawn image. They enhanced the PSGN~\cite{DBLP:journals/corr/FanSG16} method with a viewpoint estimation module. To alleviate the deformation of sketches, they proposed a sketch standardization module to alleviate the deformation problem of sketches. The work~\cite{zhong2020deep} discussed the additional challenges of line drawings in comparison with images in 3D reconstruction. In \cite{zhong2020towards}, sketches from two viewpoints were used as the input to perform 3D reconstruction, and they collected two datasets, ProSketch and AmateurSketch. Sketch2model~\cite{zhang2021sketch2model} alleviated the ambiguity in sketch modeling by decoupling view code and shape code. Sketch2mesh~\cite{guillard2021sketch2mesh} used an encoder/decoder architecture to learn a latent representation of an input sketch and refined it by matching the external contours of the reconstructed 3D mesh to the sketch during the inference process. While achieving good performance, this approach is time-consuming. Most deep sketch modeling methods encode a sketch as a latent code and then apply a decoder to convert the latent code to a 3D shape. However, these approaches fail to preserve spatial details in a sketch.

\subsection{3D Reconstruction from Single RGB Image}
3D reconstruction is a problem that has been widely studied in computer vision. Reconstructing a 3D shape from a single image is an ill-posed problem that requires strong prior knowledge. In recent years, with the development of deep learning, neural networks can be used to extract useful features for 3D reconstruction~\cite{xie2019pix2vox,ijcv/XieYZZS20,eccv/PopovBF20,DBLP:journals/corr/FanSG16}. The early works focus on reconstructing 3D shapes represented by regular voxels~\cite{choy20163d,wu2017marrnet,wang20173densinet}. MarrNet~\cite{wu2017marrnet} and 3DensiNet~\cite{wang20173densinet} resorted to intermediate representations (i.e., 2.5D sketches and density heat-map) to facilitate reconstruction. In this work, we introduce the density map as a proxy, which reflects the probability of points projected at each location of the image plane.
Unlike MarrNet and 3DensiNet that directly reconstruct 3D shapes based on the intermediate representations, we use the density map as the guidance for 2D points sampling, from which the depth value will be further predicted.

However, voxel reconstruction is inefficient because the information of 3D shape is distributed only on the surface of an object. Therefore, some methods~\cite{DBLP:journals/corr/FanSG16,wang2018pixel2mesh} attempted to recover the surface information of 3D shapes, such as point clouds or meshes.
Nevertheless, the surface of a 3D shape is sparse and irregular,
posing great challenges for shape recovery. Some works~\cite{wang2018pixel2mesh,gkioxari2019mesh} use the coarse-to-fine and feature pooling strategies to alleviate this problem. In addition to reconstructing a 3D shape based on explicit representations, recent approaches~\cite{xu2019disn,mescheder2019occupancy,bmvc/BianWLP21} explore 3D reconstruction based on implicit surface learning. But these methods require post-processing to obtain explicit 3D shapes. 
\section{Methodology}
\begin{figure*}[t]
  \centering
  \includegraphics[width=\linewidth]{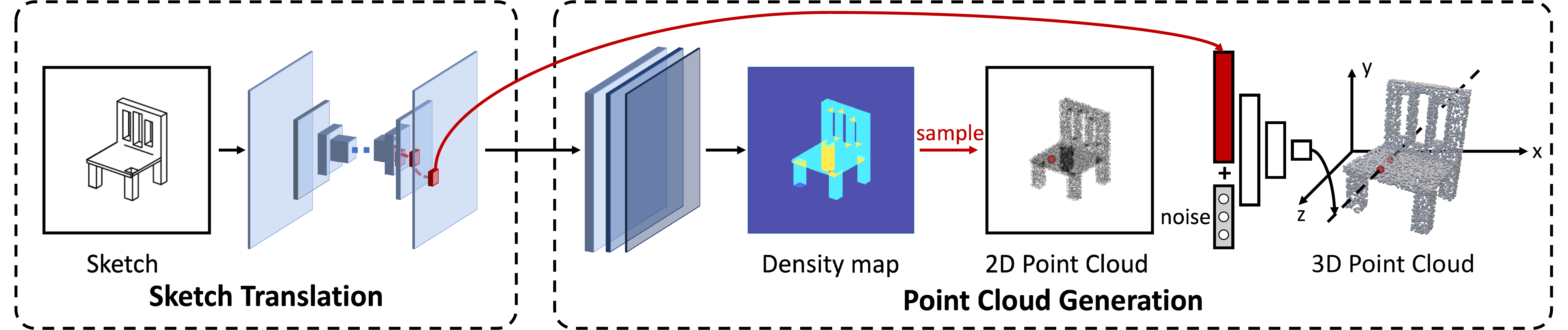}
  \caption{
    Overview of our proposed method. It consists of sketch translation and point cloud generation. During sketch translation, missing information such as surface and occlusions are implicitly compensated. The resultant feature maps are used to predict the density map. When generating 3D points, a 2D point cloud is first sampled based on the density map, followed by sampling the depth values at each given 2D point. 
  }\label{fig:architecture}
\end{figure*}

As shown in Fig.~\ref{fig:architecture}, our sketch-based modeling framework mainly consists of two components: a sketch translator and a point cloud generator. Given an input sketch $I$, the \textbf{sketch translator} first translates it to a feature map $F$. Next, the \textbf{point cloud generator} produces a point cloud $\mathcal{S}$ based on the given feature map $F$. When recovering the point cloud, a 2D density map is first predicted, from which 2D points are sampled; then the depth of each 2D point is predicted by using the proposed conditional depth generator. Note that in line with \cite{guillard2021sketch2mesh,nguyen2019graphx}, we adopt the commonly used “viewer-centered" setting \cite{shin2018pixels}, in which we assume the image space and the 3D space are aligned.

\subsection{Sketch Translation}
\label{subsec:stm}
The goal of the sketch translator is to fully exploit the spatial information in a sketch and generate suitable features for 3D shape prediction. 
It is non-trivial because there is a large information discrepancy between a sketch and a 3D shape: 1) a sketch is sparse and mainly preserves structural framework of a corresponding 3D shape. The object surface, occluded object surface, and vacancy between surfaces can all be shown as blank in a sketch.
2) most depth information in a 3D shape is also lost in a sketch image. Therefore, the sketch translator aims to complement the missing information. 
For example, inferring whether a blank area belongs to an object, or which pixels are on the same surface. 

Specifically, we adopt an encoder-decoder based CNN network for sketch translation. Firstly, an encoder network is used to extract features from the input sketch with multiple down-sampling blocks. This is to increase the receptive field of the neurons to acquire an overview of the input sketch. A decoder network consisting of multiple upsampling blocks is then used to gradually infer the information of the 3D shape with increased spatial resolution. Instead of using the last feature map $F^{n}$ for point cloud generation, we use a similar idea~\cite{lin2017feature} that leverages the feature maps at all scales by upsampling the feature map at each individual scale $F^{i}$ to the size of $F^{n}$ and concatenating them together to produce the final feature $F$.

After sketch translation, the response of the feature map $F$ is much denser than the input sketch while the spatial alignment and resolution are roughly maintained. 
It will facilitate the prediction of point clouds of with fine details, which will be explained below.
 
 \subsection{Point Cloud Generation}
 
The point cloud generator aims to recover the point cloud of the corresponding 3D shape $\mathcal{S}$ from the translated feature maps $F$. To utilize the spatial information of the input sketch, we decompose the point could generation process into two steps: 1) predicting the 2D point cloud which is the projection of the 3D point cloud into the image plane; 2) inferring the depth of each 2D point. 

For generating a 2D point cloud, the point cloud generator predicts the joint distribution of the coordinates from the projected points $p(X, Y|I)$, where $X,Y$ are random variables corresponding to the $x, y$ axis respectively. Sampling from $P(X, Y|I)$ will generate the 2D point cloud. Fig.~\ref{fig:motivation} shows an example of projecting a 3D shape into the image plane and its corresponding density map. The probabilistic density at each location varies because it depends on how many surfaces are being passed by the ray centered at this location. 

After a 2D point cloud is generated, the point cloud generator predicts the depth distribution of each point $p(Z_i|x_i,y_i,I)$, where $x_i, y_i$ is the location of the $i$-th point in the image plane. Sampling from $P(Z_i|x_i,y_i,I)$ gives the depth of each point. Combining $x, y$ coordinates from density map sampling and $z$ coordinate from depth sampling, the overall 3D point cloud can be generated.

From a probabilistic view, this process actually models the shape of a 3D object as a joint distribution of $x, y, z$ coordinates.
Our generation process assumes a factorization process over projection and conditional independency between different locations for depth prediction, i.e., $P(X,Y,Z|I)=P(X, Y|I)P(Z|X,Y,I)$. The first term and the second respectively correspond to the process of generating a 2D point cloud and the process of predicting depth given a 2D point cloud and a sketch.

\subsubsection{2D Point Cloud Generation.}\quad
As all valid locations must lay inside the image, we firstly model the distribution of projected points in pixel coordinates. The image coordinates can be seen as quantizing the $x, y$ location into $\mathcal{W}\times \mathcal{H}$ bins, where $P_{u,v}=P(X=u, Y=v)$ is the probability of a projected point inside the $(u, v)$-th bin. 

We use a mask prediction head to directly predict the density map $M \in \mathcal{R}^{\mathcal{ W \times H }}$, where $M_{v,u}=P_{u,v}$. It takes the translated sketch feature $F$ as the input, and resizes the feature map to the size of ${\mathcal{ W \times H }}$ by using bilinear interpolation. The interpolated feature map is then passed to three convolutional layers for density prediction. The hyper parameters $\mathcal{W}$ and $\mathcal{H}$ control the resolution of the point clouds.

To generate a 2D point cloud, we can see $P(X,Y)$ as a multinomial distribution over $\mathcal{W}\times \mathcal{H}$ locations.
We firstly sample a specific number of locations with the probabilities defined by the density map $M$, and then use the column and row indices $u,v$ as $x, y$ coordinates. We normalize the coordinate to the range of $[-1, 1]$ \footnote{$x=\frac{2u}{\mathcal{W}-1}-1$, $y=\frac{2v}{\mathcal{W}-1}-1$, where $u=0,1,...,\mathcal{W}-1$, $v=0,1,...,\mathcal{H}-1$} to produce the coordinate in the image plane $(x^I,y^I)$. It then converts the points to world coordinates by using the camera parameters. As we use the orthogonal projection model to produce the rendered sketches, the $x,y$ coordinates are linearly mapped from that of the 3D point clouds. That is the $x,y$ coordinates of a point in the raw 3D point cloud, and $(x,y)$ can be computed as  $(x^I/s, y^I/s)$, where $s$ is a preset parameter of the projection model. Similar mapping functions can be drawn for other projection models.

\subsubsection{Conditional Depth Estimation.}\quad
After producing the 2D coordinates $(x,y)$ of the 3D points, the next step is to predict their $z$ coordinates. Given its $x,y$ location, we assume estimating the depth for each individual point to be independent, so we predict the conditional depth distribution $P(Z_i|x_i, y_i)$ separately for each 2D location. Given a $x,y$ location, the depth distribution $P(Z_i|x_i, y_i)$ can be multimodal and the number of modes tends to be varied, as there may be one or multiple points from different surfaces sharing the same 2D location. It is hard to explicitly define the probabilistic function of $P(Z_i|x_i, y_i)$.

Inspired by the Generative Adversarial Networks~\cite{goodfellow2014generative,mirza2014conditional,isola2017image}, we use an implicit approach and adopt the generator network design to model $P(Z_i|x_i, y_i)$. It takes a noise variable $N \in R^{d}$ and the local feature $f_{x,y}$ as input, and predict a scalar of depth $z$, where $N$ is sampled from the uniform distribution $\mathcal{U}(0,1)$ 
and $f_{x,y}$ is obtained by extracting from the feature map $F$ at the corresponding location. Note that the depth generator can output different depth values given the same feature and different noise variables. 
It takes a multi-layer perceptron (MLP) as the backbone and its parameters are shared at all $(x_i,y_i)$ locations. 

For inference, we randomly sample a noise vector $\mathbf{n}_i$ by following the uniform distribution for each point $(x_i,y_i)$ in the predicted 2D point cloud, and then predict the corresponding $z_i$. Putting the 2D location $(x_i,y_i)$ and depth prediction $z_i$ together will generate the final point cloud $\mathcal{S}$. Note that the sampled random noise $\mathbf{n}_i$ controls which mode the predicted depth $z_i$ falls in if the corresponding $P(Z_i|x_i,y_i,I)$ is multimodal. Together with 2D point cloud sampling, the two-stage process can be seen as sampling from the joint distribution that defines the coordinates of a 3D shape.
The detailed process of point cloud generation is listed in Algorithm~\ref{algo}. Note that our proposed method is compatible with both orthogonal projection and perspective projection. It can be controlled by the `invproj' function in Algorithm~\ref{algo}.
\renewcommand{\algorithmicrequire}{\textbf{Input:}}
\renewcommand{\algorithmicensure}{\textbf{Output:}}
\begin{algorithm}[t]
\caption{Point Cloud Generation Process}\label{alg:cap}
\begin{algorithmic}[1]
\Require
total number of points $N$, the predicted density map $M$, feature map $F$, and  depth generator $T$.
\Ensure
the reconstructed point clouds of the sketch $\mathcal{S}$.
\State Let $\mathcal{S} = \emptyset $
\While{$|\mathcal{S}| \leq N$}
    \State sample a location from the multinomial distribution defined by $M$, i.e. $(u,v) \sim Mult(x,y;M)$.
    \State convert $u,v$ to the image plane coordinate $x^I,y^I$ 
    \State sample the noise vector $\mathbf{n} \sim \mathcal{U}(0,1)$.
    \State inference the depth at $u,v$: $z_{c}=T(\mathbf{n}, F_{uv})$
    \State convert $(x^I, y^I, z_{c})$ to the world coordinate: $(x,y,z)={\rm invproj}(x^I, y^I, z_{c})$
    \State $\mathcal{S}=\mathcal{S} \cup \{(x,y,z)\}$
\EndWhile \\
\Return $\mathcal{S}$
\end{algorithmic}
\label{algo}
\end{algorithm}
 
\subsection{Loss Function}
  
A key role in our proposed approach is the density map. Fortunately, we can freely produce the ground-truth density map from a 3D shape by a customized renderer, i.e., counting the number of points that occurred when projecting a ray from a 3D point onto an image plane followed by normalization. To provide supervision information for the learning process of the density map, we use the L1 loss as a constraint, as shown in Eq.~\eqref{eq:density}.
\begin{equation}
   L_{D}=\sum_{x_i,y_i} \|\hat{p}(x_i,y_i) - p(x_i,y_i)\|_{1}. 
\label{eq:density}
\end{equation}
To provide supervision information for the learning process of the conditional generator, we constrain the distance between the output point cloud and the ground-truth point cloud. We use the Chamfer distance as the loss function during the training process. Given two point clouds $\mathcal{S}, \mathcal{\hat{S}} \subseteq \mathcal{R}^{3}$, the Chamfer distance is defined as Eq.~\eqref{eq:cd}. The final loss function is shown in Eq.~\eqref{eq:total}, and $\lambda_{1}$ and $\lambda_{2}$ are the weights of $L_{CD}$ and $L_{D}$, respectively.
\begin{equation}
    L_{CD}=\frac{1}{|\mathcal{S}|} \sum_{p \in \mathcal{S}} \min _{q \in \mathcal{\hat{S}}}\|p-q\|_{2}^{2}+\frac{1}{|\mathcal{\hat{S}}|} \sum_{q \in \mathcal{\hat{S}}} \min _{p \in \mathcal{S}}\|q-p\|_{2}^{2}
    \label{eq:cd}
\end{equation}
\begin{equation}
L = \lambda_{1} L_{CD} + \lambda_{2} L_{D},
\label{eq:total}
\end{equation}

As shown in Fig.~\ref{fig:architecture}, during the training process, the feature maps from the encoder-decoder network are fed into two paths: 1) the convolutional layers to predict the density map; 2) the fully-connected layers to predict the depth value. Correspondingly, the L1 loss in Eq.~\eqref{eq:density} and the Chamfer loss in Eq.~\eqref{eq:cd} are computed, and the gradients from these two losses will be separately backpropagated along two different paths back to the encoder-decoder network.
  
\section{Experiment}
In this section, we first introduce the datasets and evaluation metrics used in our experiments and provide implementation details (Sec.~\ref{sec:datasets}). We then compare our proposed method with the baseline methods on the Synthetic-LineDrawing dataset (Sec.~\ref{sec:quantitative_results}) and three hand-drawn sketch datasets (Sec.~\ref{sec:fhs}). Ablation studies are conducted to show the effectiveness of individual modules (Sec.~\ref{sec:ablation}).
We also evaluate our model on unseen classes (Sec.~\ref{sec:scalability}).

\subsection{Experimental Setup and Evaluation Metrics}
\label{sec:datasets}

\noindent \textbf{Synthetic-LineDrawing dataset.} \quad Publicly available large-scale paired sketch-3D datasets are rare. So we contribute a new dataset, the Synthetic-LineDrawing dataset, by rendering sketch images from 3D models of the ShapeNet dataset~\cite{shapenet2015}. Specifically, we use a subset of ShapeNet-core, which consists of around 50k 3D models spanning 13 classes. We select 5 random views for each object to render, resulting in 218,915 sketch images and corresponding 43,783 3D objects.
We follow the conventional train/test splits as in \cite{choy20163d}, i.e., $4/5$ and $1/5$ for training and test, respectively. 

Except the synthetic sketch dataset, we also conduct experiments on three hand-drawn sketch datasets:
\begin{itemize}
    \item \textbf{ShapeNet-Sketch}~\cite{zhang2021sketch2model} is a dataset consisting of $1,300$ free-hand sketches and their corresponding ground-truth 3D models, belonging to the same 13 categories of the ShapeNet dataset. All sketch images are drawn by volunteers with different drawing skills.
    \item \textbf{AmateurSketch}~\cite{zhong2020towards} contains $3,015$ sketch images of $500$ chair models and 1,665 sketch images of 555 lamp models. Each 3D model is drawn from 3 different viewpoints.
    \item \textbf{ProSketch-3DChair}~\cite{zhong2020towards} contains $1,500$ professional sketches of $500$ chair models, and each 3D model is drawn from 3 different viewpoints: front, side and 45 degree.
\end{itemize}

\subsubsection{Implementation Details.}\quad
We use a CNN-based encoder-decoder network~\cite{isola2017image} as the sketch translator. For density map prediction, we use ReLU followed by normalization to ensure the sum of the values from all spatial positions of the density map is 1. When estimating the depth information, an MLP with residual connections is used. $\lambda_1$ and $\lambda_2$ are set to be 1 and $10^4$, respectively. The model is trained for 30 epochs with an initial learning rate of $10^{-3}$. Adam optimizer~\cite{kingma2014adam} is used for optimization. The ground-truth density map is obtained by counting the number of points that occurred when a ray projects from a 3D point onto the image plane. 

\subsubsection{Evaluation Metrics.} \quad We employ four evaluation metrics to measure the reconstruction performance on the above four datasets: Chamfer Distance (CD), Earth Mover’s Distance (EMD), voxel Intersection over Union (Vox-IoU), and Fréchet Point cloud Distance (FPD). \textbf{CD} is a widely used as the evaluation metric for 3D generation and reconstruction tasks. Similar to CD, \textbf{EMD} is also used to evaluate the similarity between two point clouds. But it is more 
sensitive to the local details and density distribution. \textbf{FPD} is similar to FID, which calculates the 2-Wasserstein distance between the real and fake samples in the feature space extracted by a pre-trained PointNet. \textbf{Voxel-IoU} measures the coverage percentage of two volumetric models.
Further details of these evaluation metrics are explained in Supplementary.

\subsection{Results on Synthetic-LineDrawing Dataset}
\label{sec:quantitative_results}

\subsubsection{Baseline methods.}\quad We first compare our approach with three state-of-the-art methods for \textit{sketch-based} single-view 3D reconstruction (SVR):

\textbf{Sketch2Mesh}~\cite{guillard2021sketch2mesh}: Given an input sketch, this method also utilizes an encoder-decoder network to produce a 3D mesh estimate. It learns a compact feature representation and recovers the 3D shape by minimizing the 2D Chamfer distance between the 3D shape's projected contour and the input sketch.

\textbf{Sketch2Model}~\cite{zhang2021sketch2model}: This method is proposed to reconstruct a 3D shape represented by a mesh. It employs an encoder-decoder network. To address the ambiguity problem of sketch, it introduces an additional encoder-decoder to decompose a sketch image to the view and shape space. During the inference process, each 3D shape is reconstructed based on an input sketch and the estimated viewpoint.

\textbf{Sketch2Point}~\cite{wang20203d}: This method is proposed to reconstruct a 3D point cloud from a sketch. It is built on PSGN~\cite{DBLP:journals/corr/FanSG16}, in which the key component is a standardization module used to handle sketches with various drawing styles.

\newcommand{\addFig}[1]{{\includegraphics[height=.092\textwidth]{#1.png}}}
\begin{figure*}[t]
  \centering
  \setlength{\tabcolsep}{1mm}
  \begin{tabular}{cccccccc}
   \addFig{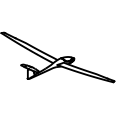}&
    \addFig{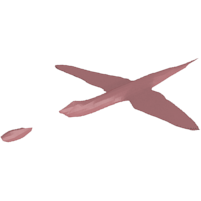}&
   \addFig{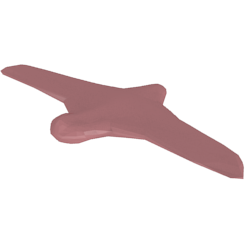}&
    \addFig{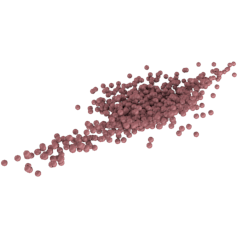}&
   \addFig{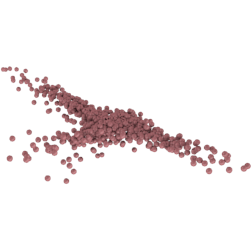}&
   \addFig{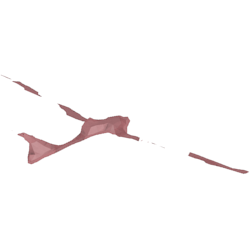}&
   \addFig{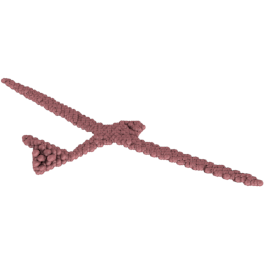}&
   \addFig{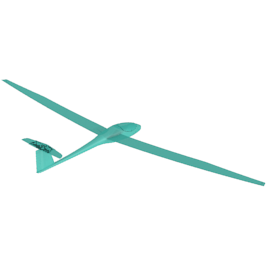}
   \\
   \addFig{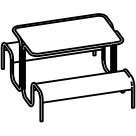}&
    \addFig{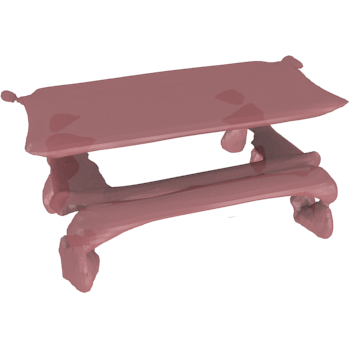}&
   \addFig{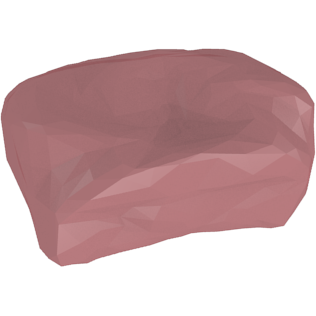}&
    \addFig{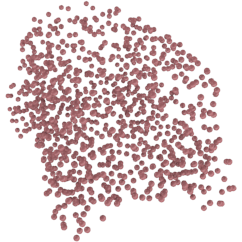}&
   \addFig{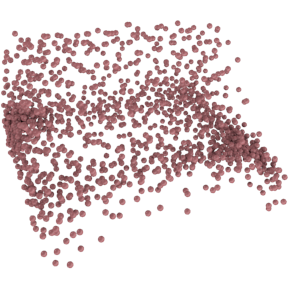}&
   \addFig{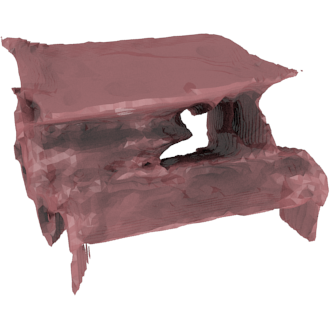}&
   \addFig{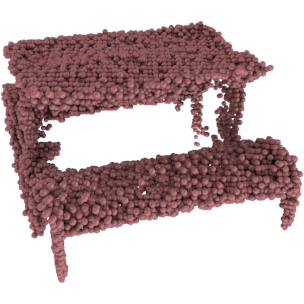}&
   \addFig{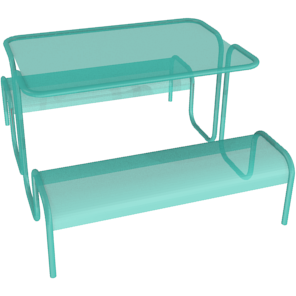}
   \\
   \addFig{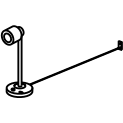}&
    \addFig{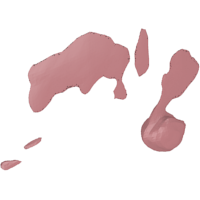}&
   \addFig{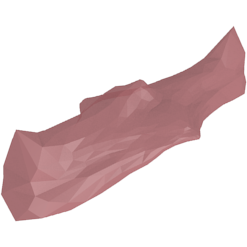}&
    \addFig{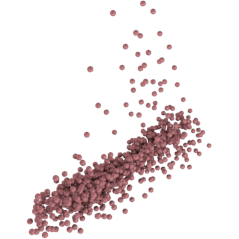}&
   \addFig{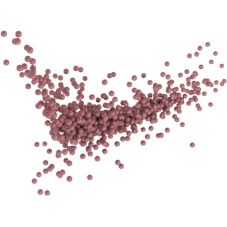}&
   \addFig{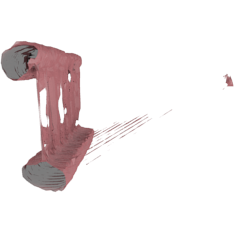}&
   \addFig{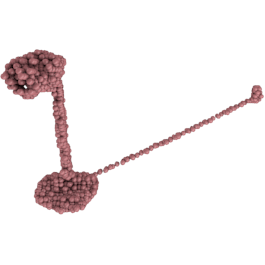}&
   \addFig{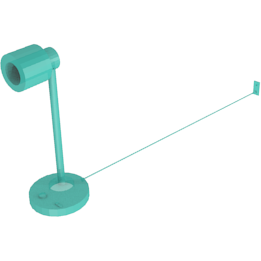}
   \\
   \addFig{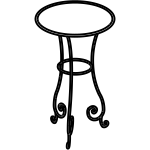}&
    \addFig{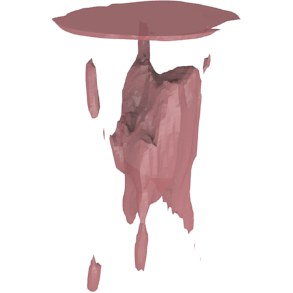}&
   \addFig{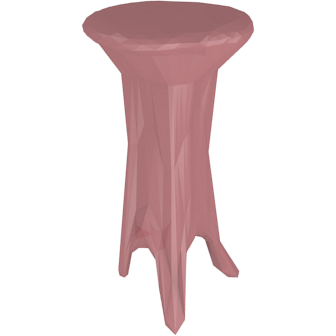}&
    \addFig{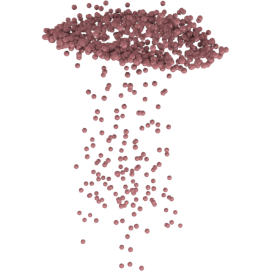}&
   \addFig{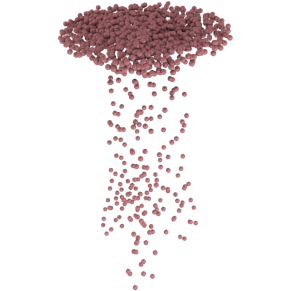}&
   \addFig{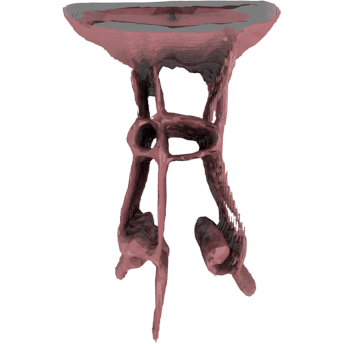}&
   \addFig{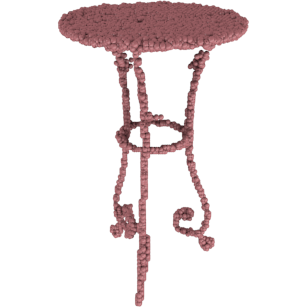}&
   \addFig{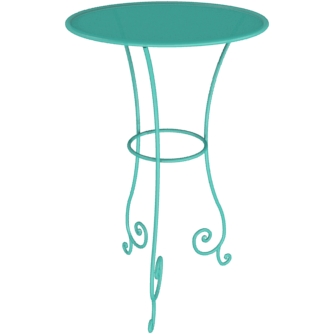}
   \\
   sketch & 
   \cite{guillard2021sketch2mesh} 
   & 
   \cite{zhang2021sketch2model} 
   & 
   \cite{wang20203d} 
   & 
   \cite{nguyen2019graphx} 
   & 
   \cite{xu2019disn} 
   & Ours & GT\\
  \end{tabular}
  \caption{
    Reconstruction results on the Synthetic-LineDrawing Dataset.
    }
  \label{fig:res_main}
\end{figure*}
\renewcommand{\addFig}[1]{{\includegraphics[height=.08\textwidth]{#1.png}}}
\begin{figure*}[t]
  \centering
  \setlength{\tabcolsep}{0.5mm}
  \begin{tabular}{c|ccccccccc}
   \addFig{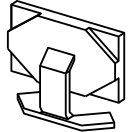}&
   &
   \addFig{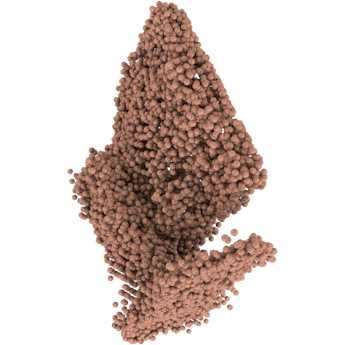}&
   \addFig{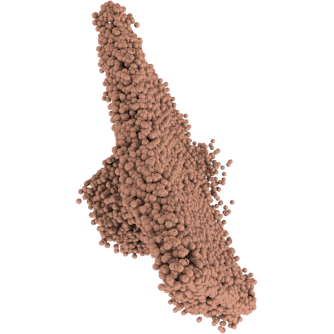}&
   \addFig{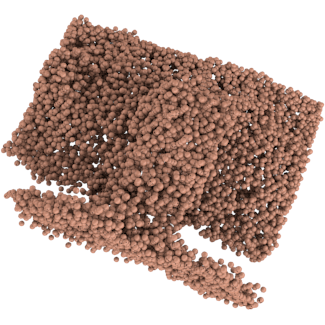}&
   \addFig{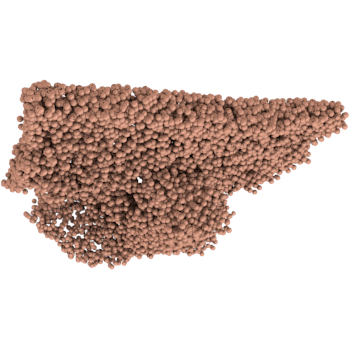}&
   \addFig{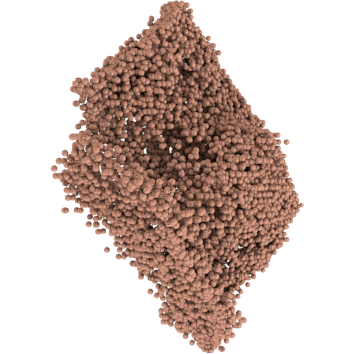}&
   \addFig{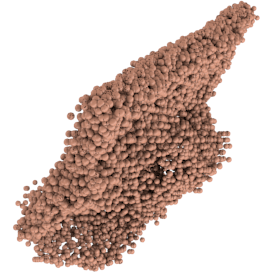}&
   \addFig{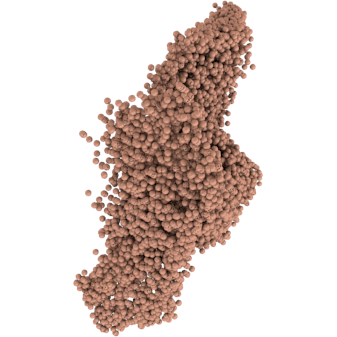}&
   \addFig{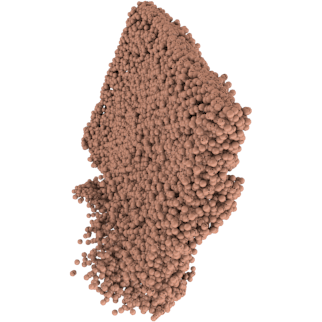}
   \\
   \addFig{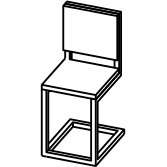}&
   &
   \addFig{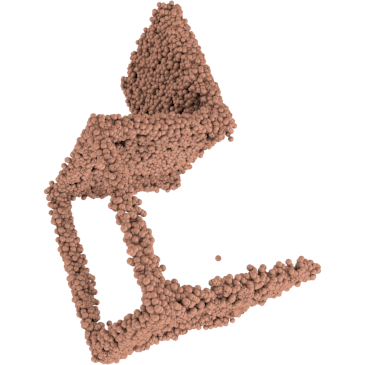}&
   \addFig{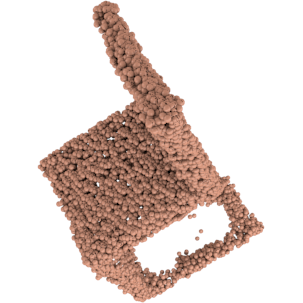}&
   \addFig{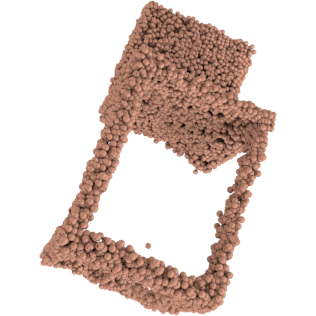}&
   \addFig{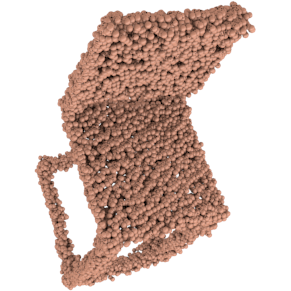}&
   \addFig{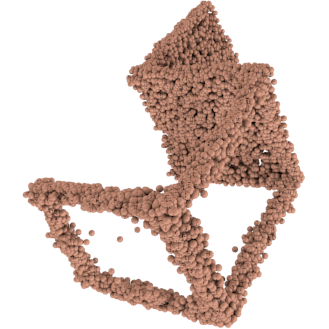}&
   \addFig{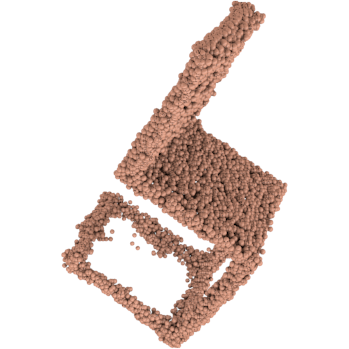}&
   \addFig{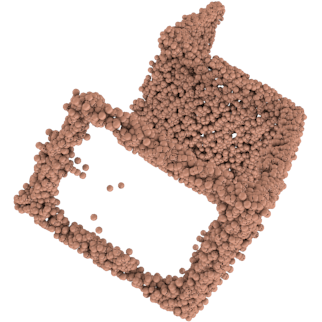}&
   \addFig{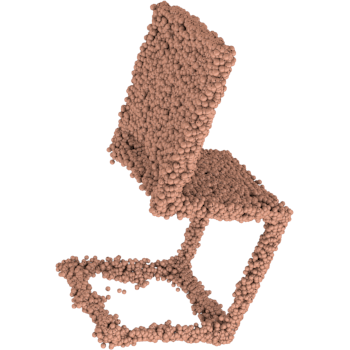}
  \\
   sketch & view & \#1 & \#2 & \#3 & \#4 & \#5 & \#6 & \#7 & \#8\\
  \end{tabular}
  \caption{
    Our reconstructed 3D shapes that are rendered under different viewpoints. The results show that the generated 3D point clouds are consistently accurate and faithful under all viewpoints.}
  \label{fig:res_views}
\end{figure*}

\begin{table*}[t]
  \centering
  \caption{
    Results on the Synthetic-LineDrawing Dataset.
  }
  \resizebox{\textwidth}{!}{
  \begin{tabular}{lcccccccccccccc} \toprule
              &\multicolumn{13}{c}{ Categories}&\multirow{2}{*}{ \textbf{mean}} \\ \cline{2-14}
               & airplane & bench & cabinet & car & chair & display & lamp & speaker & rifle & sofa & table & phone & boat & \\  
               \midrule
               \multicolumn{15}{c}{ Chamfer Distance($\downarrow$) $\times10^{-3}$}
               \\\midrule
    Sketch2Mesh\cite{guillard2021sketch2mesh} & 0.910 & 4.533 & 2.735 & 1.417 & 3.002 & 3.119 & 9.054 & 4.685 & 0.846 & 2.633 & 2.732 & 2.005 & 2.524 & 3.092 \\
    Sketch2Model\cite{zhang2021sketch2model}  & 1.814 & 6.404 & 3.010 & 2.720 & 3.997 & 6.976 & 6.617 & 5.579 & 1.495 & 5.721 & 4.632 & 2.723 & 2.755 & 4.188 \\
    Sketch2Point\cite{wang20203d}   & 2.229 & 14.747 & 3.239 & 1.610 & 3.883 & 7.047 & 6.663 & 6.611 & 4.056 & 7.132 & 5.772 & 4.392 & 1.255 & 5.280 \\
    PCDNet\cite{nguyen2019graphx}  & 0.571 & 1.151 & 1.480 & 1.002 & 1.664 & 1.389 & 3.104 & 2.120 & 0.621 & 1.416 & 1.647 & 1.207 & 1.051 & 1.417 \\
    DISN\cite{xu2019disn}  & 0.845 & 3.573 & 1.839 & 1.340 & 3.181 & 2.640 & 9.203 & 3.340 & 2.000 & 1.797 & 3.371 & 2.080 & 2.215 & 2.879 \\
    \textbf{Ours}  & \textbf{0.389} & \textbf{0.729} & \textbf{1.153} & \textbf{0.866} & \textbf{1.033} & \textbf{0.959} & \textbf{1.907} & \textbf{1.561} & \textbf{0.428} & \textbf{1.050} & \textbf{0.949} & \textbf{0.957} & \textbf{0.780} & \textbf{0.982} \\\midrule
               \multicolumn{15}{c}{Earth Mover's Distance($\downarrow$) $\times10^{-2}$}
               \\\midrule
    Sketch2Mesh\cite{guillard2021sketch2mesh}  & 3.914 & 5.732 & 5.441 & 4.645 & 6.032 & 5.301 & 11.188 & 7.005 & 5.113 & 5.297 & 5.328 & 4.133 & 4.804 & 5.687 \\
    Sketch2Model\cite{zhang2021sketch2model}   & 5.587 & 7.460 & 5.852 & 5.662 & 6.867 & 7.436 & 10.615 & 7.109 & 6.297 & 7.035 & 7.206 & 4.679 & 6.529 & 6.795 \\
    Sketch2Point\cite{wang20203d}  & 6.893 & 13.907 & 7.102 & 5.875 & 9.913 & 10.799 & 15.212 & 9.736 & 10.556 & 10.143 & 9.263 & 8.652 & 5.880 & 9.533 \\
    PCDNet\cite{nguyen2019graphx}  & 7.114 & 8.723 & 9.745 & 7.420 & 10.948 & 9.493 & 16.054 & 10.465 & 7.464 & 10.121 & 10.450 & 7.880 & 7.255 & 9.472 \\
    DISN\cite{xu2019disn}   & 3.823 & 6.234 & \textbf{4.911} & 4.569 & 7.136 & 5.893 & 10.469 & 6.063 & 5.513 & 4.706 & 6.990 & 4.053 & 5.037 & 5.800 \\ 
    \textbf{Ours}  & \textbf{3.178} & \textbf{3.978} & 5.032 & \textbf{4.240} & \textbf{5.293} & \textbf{4.553} & \textbf{6.722} & \textbf{5.690} & \textbf{3.436} & \textbf{4.662} & \textbf{4.421} & \textbf{3.762} & \textbf{3.969} & \textbf{4.534} \\\midrule
               \multicolumn{15}{c}{Fréchet Point Cloud Distance($\downarrow$) $\times10$}
               \\
    \midrule
    Sketch2Mesh\cite{guillard2021sketch2mesh} & 2.030 & 8.253 & 3.346 & 1.147 & 3.214 & 3.287 & 10.771 & 2.608 & 1.577 & 3.436 & 1.624 & 3.647 & 7.565 & 4.039 \\
    Sketch2Model\cite{zhang2021sketch2model} & 1.524 & 13.900 & 5.546 & 1.121 & 3.220 & 9.622 & 2.887 & 6.364 & 3.494 & 20.001 & 4.393 & 3.188 & 5.490 & 6.212 \\
    Sketch2Point\cite{wang20203d} & 11.415 & 22.056 & 6.466 & 6.973 & 25.730 & 12.369 & 6.903 & 11.431 & 27.448 & 13.391 & 21.120 & 14.425 & 3.322 & 14.081 \\
    PCDNet\cite{nguyen2019graphx} & 0.991 & 1.117 & 0.760 & 1.107 & 0.846 & 0.892 & 1.657 & 1.177 & 0.916 & 0.957 & 1.050 & 0.760 & 1.313 & 1.042 \\
    DISN\cite{xu2019disn} & 1.838 & 5.097 & 1.037 & \textbf{0.285} & 2.752 & 1.662 & 12.211 & 1.294 & 3.867 & 1.729 & 3.143 & 2.734 & 2.595 & 3.096 \\ 
    \textbf{Ours} & \textbf{0.516} & \textbf{0.542} & \textbf{0.358} & 0.427 & \textbf{0.454} & \textbf{0.519} & \textbf{1.082} & \textbf{0.734} & \textbf{0.635} & \textbf{0.561} & \textbf{0.633} & \textbf{0.347} & \textbf{0.729} & \textbf{0.580} \\\midrule
               \multicolumn{15}{c}{Voxel-IOU($\uparrow$)}
               \\
    \midrule
    Sketch2Mesh\cite{guillard2021sketch2mesh} & 0.693 & 0.506 & 0.383 & 0.515 & 0.442 & 0.469 & 0.355 & 0.280 & 0.691 & 0.418 & 0.493 & 0.596 & 0.553 & 0.492 \\
    Sketch2Model\cite{zhang2021sketch2model}  & 0.499 & 0.220 & 0.338 & 0.341 & 0.308 & 0.250 & 0.320 & 0.229 & 0.511 & 0.245 & 0.269 & 0.535 & 0.422 & 0.345 \\
    Sketch2Point\cite{wang20203d} & 0.427 & 0.174 & 0.172 & 0.335 & 0.204 & 0.231 & 0.209 & 0.125 & 0.263 & 0.184 & 0.137 & 0.293 & 0.514 & 0.251 \\
    PCDNet\cite{nguyen2019graphx}  & 0.634 & 0.506 & 0.367 & 0.502 & 0.386 & 0.478 & 0.359 & 0.307 & 0.603 & 0.395 & 0.439 & 0.572 & 0.557 & 0.470 \\
    DISN\cite{xu2019disn}  & 0.698 & 0.464 & 0.407 & 0.521 & 0.397 & 0.462 & 0.332 & 0.325 & 0.683 & 0.437 & 0.426 & 0.627 & 0.547 & 0.487 \\ 
    \textbf{Ours}  & \textbf{0.736} & \textbf{0.619} & \textbf{0.467} & \textbf{0.563} & \textbf{0.535} & \textbf{0.577} & \textbf{0.510} & \textbf{0.412} & \textbf{0.713} & \textbf{0.500} & \textbf{0.597} & \textbf{0.654} & \textbf{0.638} & \textbf{0.578} \\
    \bottomrule
  \end{tabular}
  }
  \label{tab:camp_synsketch}
\end{table*}

We also compare our method with two \textit{image-based} SVR methods:

\textbf{PCDnet}~\cite{nguyen2019graphx}: This method can generate a 3D point cloud of arbitrary size based on a single image. It extracts the global shape feature(i.e., a feature vector) from a sketch and predicts the 3D point cloud via a deformation network. 

\textbf{DISN}~\cite{xu2019disn}: This work proposes a signed distance fields (SDF) predictor, where both global and local features are used for prediction. It can produce a 3D shape with fine details since it exploits local features sampled from image feature maps. However, this approach works slowly during the inference process. 

We follow their original works of Sketch2Model and Sketch2Mesh to train an individual model for each category. See more details in Supplementary.

\subsubsection{Qualitative Results.} \quad \label{sec:qualitative_results}
The results of different methods are illustrated in Fig.~\ref{fig:res_main}. For point cloud based methods, \textbf{Sketch2Point} and \textbf{PCDNet} perform badly where the generated 3D shape can even fall into an incorrect category, e.g., the produced 3D point cloud of a plane is more like a rifle. For those whose class labels are correct, the shapes are still considerably different from the input sketches. These observations indicate that special designs are required for accurate and generalizable sketch-based SVR models. Note that the point cloud produced by PCDNet still exhibits more details than Sketch2Point, which demonstrates the benefit of using local features.  \textbf{Sketch2Mesh} and \textbf{Sketch2Model} are mesh based methods. As they are trained for each class separately, there is no confusion between different categories. The surface of Sketch2Mesh is often disconnected on sketch with thin strokes. Regularized by view constraint, the continuity of Sketch2Model becomes better, but it tends to generate over-smoothed meshes. \textbf{DISN} is a SDF based method and it could generate almost accurate 3D reconstruction results. However, limited by the resolution of 3D grids and the use of SDF, small object parts and thin lines in a sketch are missing in the reconstruction results. 

Our method outperforms all competitors. The reconstructed point clouds are correct in terms of category labels and also exhibit notable level of details, even though our model is trained only once for all categories (i.e., class-agnostic). The overall layout of the point clouds are well aligned with the input sketches. Even small parts, e.g., the electric wire and the leg of the table, are depicted faithfully in the 3D point clouds. It suggests that our two-stage strategy is better in generalizing across different categories and capturing fine-details of sketch. The Reconstruction results from different views are also provided in Fig.~\ref{fig:res_views}.

\subsubsection{Quantitative Results.}
\quad
The quantitative results are shown in Table~\ref{tab:camp_synsketch} and we observe a similar trend with the qualitative results. 
Although the performance ranking of these models varies under different evaluation metrics, our method consistently performs the best in terms of all metrics. Moreover, our model even performs the best over almost all categories (except the \textit{cabinet} class based on the EMD metric and the \textit{car} class based on the FPD metric). It suggests that our reconstructed 3D shape captures both global structure and the local details. 
Notably, all methods perform considerably worse on the \textit{lamp} class than other classes, where the sketches contain many thin strokes with fine structures(see Fig.~\ref{fig:res_main}). Nevertheless, our methods still achieves reasonable results. A possible explanation is that the proposed 2D point cloud generation strategy ensures the points can be sampled even from very thin strokes, with which the 3D point cloud could be successfully generated.

\begin{table}[t]
  \centering
  \setlength{\tabcolsep}{2mm}
  \caption{
    Results on the hand-drawn sketch datasets. \small{*"ShapeNet-S." is short for ShapeNet-Sketch and "ProSketch" is short for ProSketch-3DChair.}
  }
  \resizebox{0.95\textwidth}{!}{
  \begin{tabular}{lcccccc} \toprule
              & ShapeNet-S. & ProSketch & AmateurSketch 
              & ShapeNet-S. & ProSketch & AmateurSketch\\  
               \cmidrule(r){2-4}\cmidrule(l){5-7}
               &\multicolumn{3}{c}{Chamfer Distance($\downarrow$) $\times10^{-3}$}
               &
               \multicolumn{3}{c}{Fréchet Point Cloud Distance($\downarrow$) $\times10$}
               \\\cmidrule(r){1-4}\cmidrule(l){5-7}
    Sketch2Mesh\cite{guillard2021sketch2mesh} 
    & 12.324 & 6.317 & 14.739 
    & 14.997 & 6.089 & 14.785\\
    Sketch2Model\cite{zhang2021sketch2model} 
    & 10.355 & 5.628 & 11.288
    & 14.449 & 5.464 & 14.600\\
    Sketch2Point\cite{wang20203d}   
    & 11.176 & 8.019 & 10.547
    & 35.864 & 38.991 & 24.794\\
    \textbf{Ours}  
    & \textbf{9.515} & \textbf{3.868} & \textbf{9.657}
    & \textbf{11.665} & \textbf{4.799} & \textbf{12.727}\\\cmidrule(r){1-4}\cmidrule(l){5-7}
               &\multicolumn{3}{c}{Earth Mover's Distance($\downarrow$) $\times10^{-2}$}
               &
               \multicolumn{3}{c}{Voxel-IOU($\uparrow$)}
               \\\cmidrule(r){1-4}\cmidrule(l){5-7}
    Sketch2Mesh\cite{guillard2021sketch2mesh} 
    & 9.947 & 7.921 & 13.164
    & 0.195 & 0.283 & 0.217\\
    Sketch2Model\cite{zhang2021sketch2model}  
    & \textbf{9.256} & 7.432 & 10.506
    & 0.205 & 0.244 & 0.199\\
    Sketch2Point\cite{wang20203d}   
    & 13.443 & 13.179 & 16.506
    & 0.163 & 0.185 & 0.166\\
    \textbf{Ours}  
    & 9.626 & \textbf{6.963} & \textbf{9.994}
    & \textbf{0.244} & \textbf{0.294} & \textbf{0.219}
    \\\bottomrule
  \end{tabular}
  }
  \label{tab:camp_fhs}
\end{table}
\subsection{Results on Hand-drawn Sketches}
\label{sec:fhs}
\renewcommand{\addFig}[1]{{\includegraphics[height=.06\textwidth]{#1.png}}}
\begin{figure}[t]
  \centering
  \setlength{\tabcolsep}{1mm}
  \begin{tabular}{cccccc|cccccc}
   \addFig{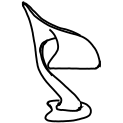}&
    \addFig{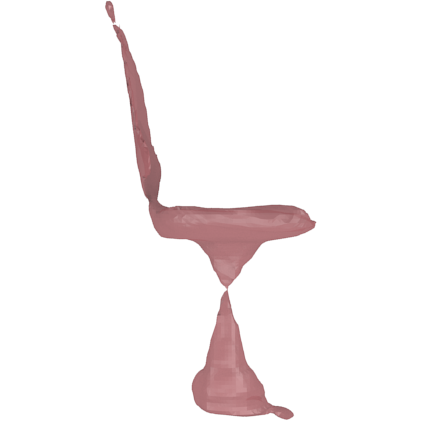}&
   \addFig{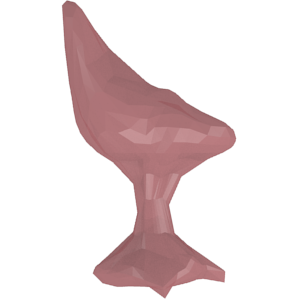}&
    \addFig{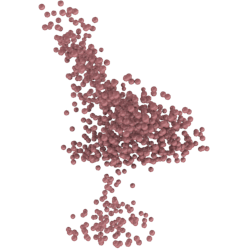}&
   \addFig{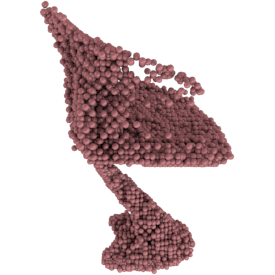}&
   \addFig{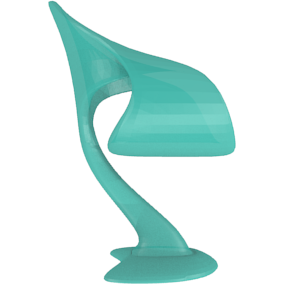}
    &
   \addFig{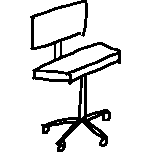}&
    \addFig{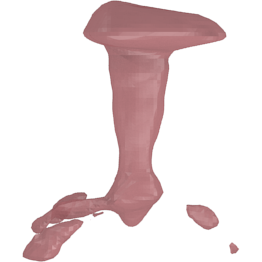}&
   \addFig{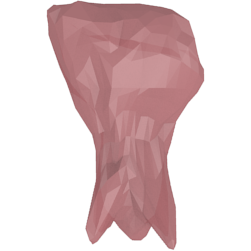}&
    \addFig{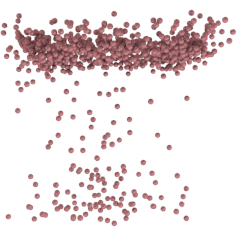}&
   \addFig{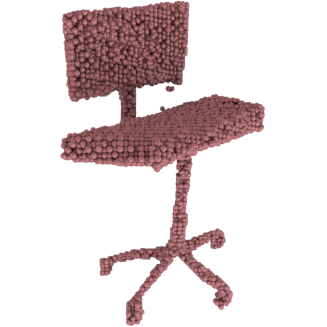}&
   \addFig{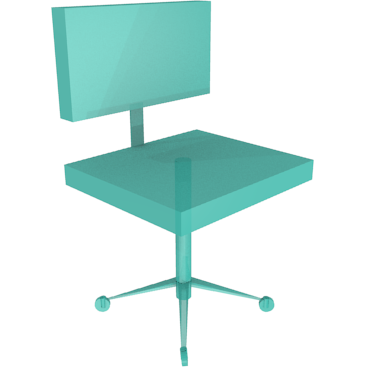}
   \\
   \addFig{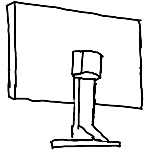}&
    \addFig{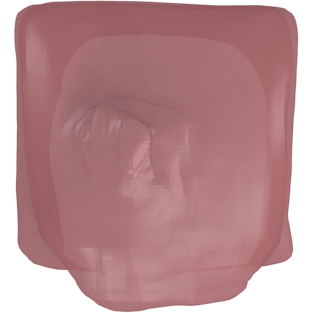}&
   \addFig{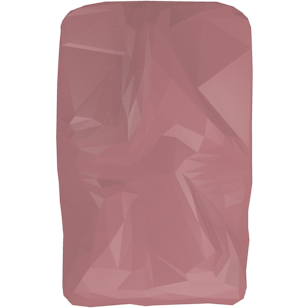}&
    \addFig{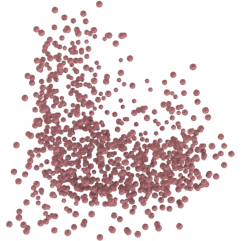}&
   \addFig{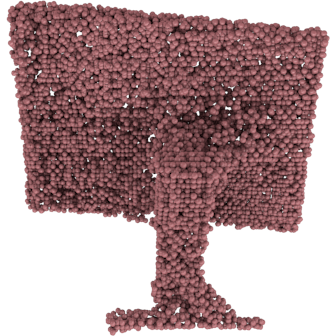}&
   \addFig{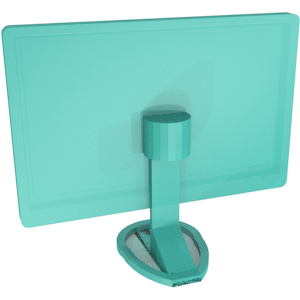}
   &
   \addFig{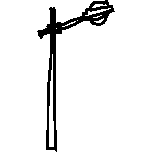}&
    \addFig{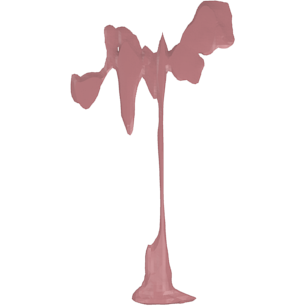}&
   \addFig{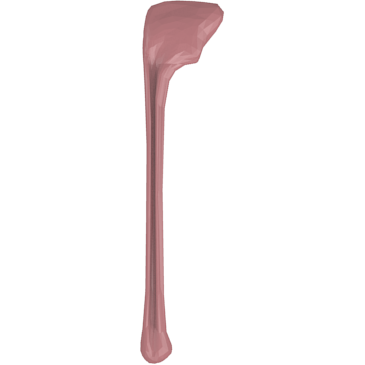}&
    \addFig{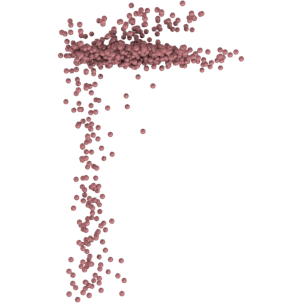}&
   \addFig{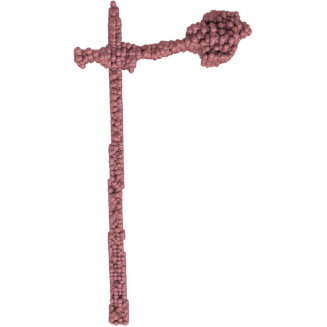}&
   \addFig{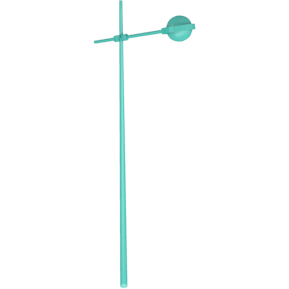}
   \\
   sketch & \cite{guillard2021sketch2mesh} & \cite{zhang2021sketch2model} & \cite{wang20203d} & Ours & GT
   &
   sketch & \cite{guillard2021sketch2mesh} & \cite{zhang2021sketch2model} & \cite{wang20203d} & Ours & GT
   \\
  \end{tabular}
  \caption{
    Reconstruction results of different methods on hand-drawn sketches.
    }
  \label{fig:res_fhs}
\end{figure}
 
In this section, we test the generalization ability of our model on three hand-drawn sketch datasets ShapeNet-Sketch~\cite{zhang2021sketch2model}, AmateurSketch~\cite{zhong2020towards}, and ProSketch-3DChair~\cite{zhong2020towards} without finetuning or domain adaptation. The three baseline methods which are specifically proposed for this task are used for comparison, including Sketch2Mesh, Sketch2Model, and Sketch2Point. The quantitative and qualitative results are shown in Table~\ref{tab:camp_fhs} and Fig.~\ref{fig:res_fhs}.
Unlike the other baseline methods where the produced 3D shape and the input sketch are aligned only at the semantic level, the 3D shapes generated by our method are much more faithful to the input sketch. However, as shown in Fig.~\ref{fig:res_fhs}, a sketch can be inaccurate. For example, the second chair's leg seems to have shifted. Our method tends to be faithful to the sketch rather than generating an object by simply following categorical shape prior. We argue that there is often a trade-off between faithfulness and rationality.
In this work, we focus on faithfulness.

\subsection{Ablation Studies}
\label{sec:ablation}
\newcommand{\addFigS}[1]{{\includegraphics[height=.10\textwidth]{#1.png}}}
\begin{figure}[t]
  \centering
  \small
  \begin{subfigure}{0.57\linewidth}
  \centering
  \begin{tabular}{cccccc}
  \addFigS{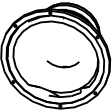}&
   \addFigS{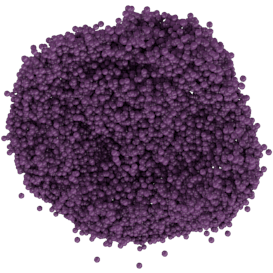}&
   \addFigS{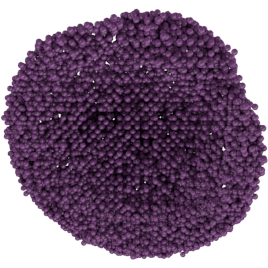}&
   \addFigS{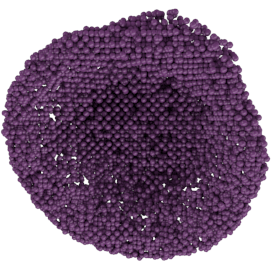}&
   \addFigS{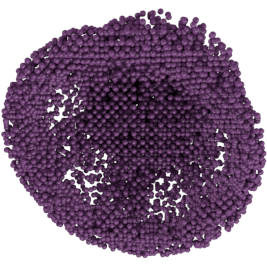}&
   \addFigS{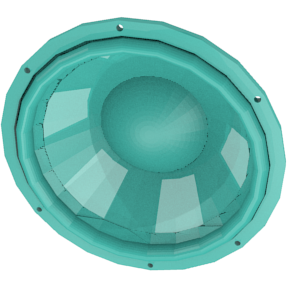}
   \\
   \addFigS{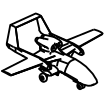}&
   \addFigS{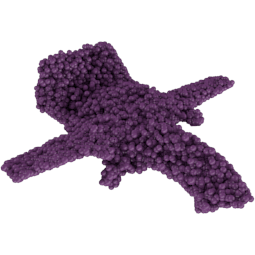}&
   \addFigS{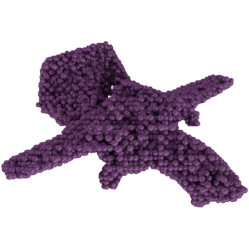}&
   \addFigS{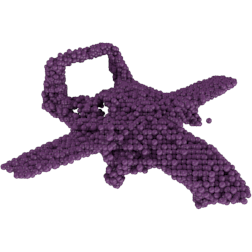}&
   \addFigS{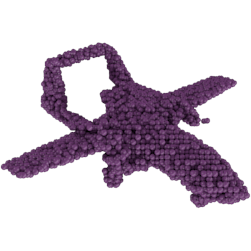}&
   \addFigS{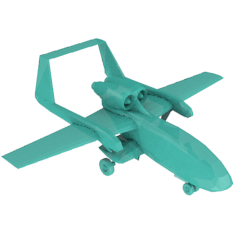}
   \\
   sketch & SimEnc & Homo & Ours & ${\rm Ours}_{real}$ & GT\\
  \end{tabular}
  \caption{}
  \label{fig:fig_abl}
  \end{subfigure}
  \begin{subfigure}{0.42\linewidth}
  \centering
  \resizebox{\textwidth}{!}{
  \begin{tabular}{lcccc}\toprule
              & CD($\downarrow$) & EMD($\downarrow$) & FPD($\downarrow$) & \multirow{2}{*}{Vox-IOU($\uparrow$)}\\ 
              &$\times10^{-3}$&$\times10^{-2}$&$\times10$& \\\midrule
    SimEnc & 1.193   & 4.779   & 1.075  &0.534   \\
    Homo & 1.106   & 5.021   & 0.841  &0.551   \\
    Ours  & 0.982   & 4.534  & 0.580 & 0.578     \\
    ${\rm Ours}_{real}$ & 0.907 & 4.194 & 0.562 & 0.591 \\
  \bottomrule
  \end{tabular}
  }
  \caption{}
  \label{tab:tab_abl}
  \end{subfigure}
  \caption{Reconstruction results of our method and different variants. ${\rm Ours}_{real}$ is a variant method which uses the ground-truth density maps.
    }
    \label{fig:abl_all}
\end{figure}

\subsubsection{Effectiveness of Sketch Translator.}\quad We use an encoder-decoder structure which is widely used for image translation to complement the information of an input sketch. Thanks to the translation module, we can produce a more informative feature map, so that a better 3D prediction can be achieved. To validate this design, we propose a comparative baseline model Simple-Encoder (`SimEnc'). When compared to our model, SimEnc removes the sketch decoder module. For a fair comparison, we increase the number of parameters of the depth estimator in SimEnc accordingly so that the amount of parameters in SimEnc is roughly the same as ours. As shown in Fig.~\ref{fig:abl_all}, we can see the performance of SimEnc drops significantly. We suppose that the feature map produced by our sketch translation module is more informative. Using the encoder network, we cannot preserve much information to help predict reasonable 3D shape.

\subsubsection{Effectiveness of Density-guided Sampler.}\quad During point generation, our method uses a density map as guidance for sampling $x,y$ coordinates. Taking into account that the inhomogeneous distribution of 3D information in 2D space can make modal transformation more effective, we compare our sampler with an alternative one denoted as homo-sampler, which treats 3D information as homogeneous distribution in two dimensions. Specifically, the homo-sampler only distinguishes between foreground and background, in which the same number of points is sampled at each position in the foreground. We use the points with $p(x,y)$ predicted by our method greater than 0 as the foreground for the homo-sampler method. The experimental results are shown in Fig.~\ref{fig:abl_all}(see the second row `Homo'). A homogeneous sampling strategy does not allow the sampler to perceive the difference in the
distribution of $p(Z_{xy} \mid X,Y)$ at different locations. We can see that the reconstruction performance when using a homo-sampler is worse than ours, as shown in Fig.~\ref{fig:abl_all}.

\renewcommand{\addFig}[1]{{\includegraphics[height=0.062\textwidth]{#1.png}}}
\begin{figure}[t]
\centering
  \setlength{\tabcolsep}{1mm}
  \begin{tabular}{cccccc|cccccc}
   \addFig{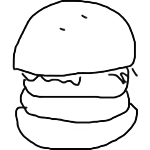}&
    \addFig{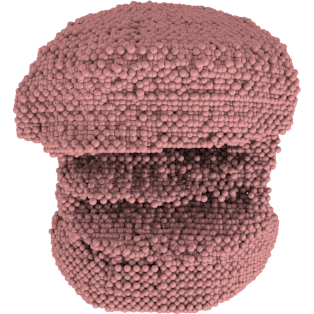}&
   \addFig{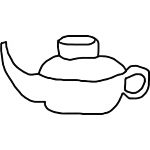}&
    \addFig{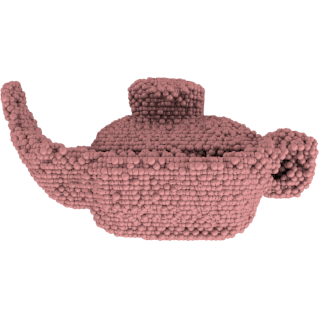}&
   \addFig{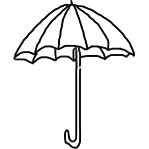}&
    \addFig{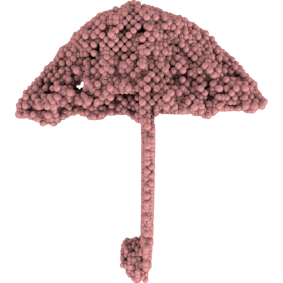}&
   \addFig{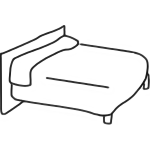}&
    \addFig{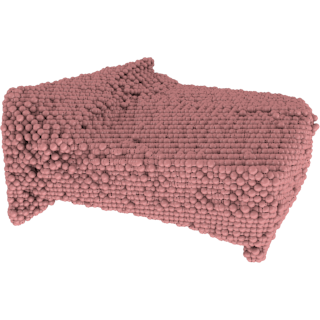}&
   \addFig{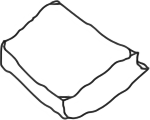}&
    \addFig{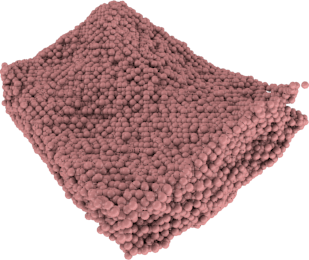}&
   \addFig{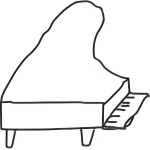}&
    \addFig{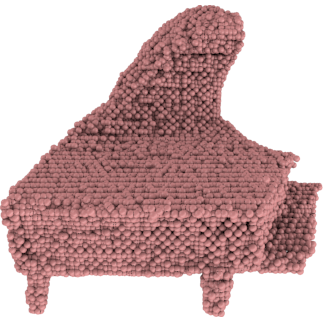}
  \end{tabular}
  \caption{Our reconstruction results on unseen classes.}
\label{fig:res_unseen}
\end{figure}
\renewcommand{\addFig}[1]{{\includegraphics[height=.063\textwidth]{#1.png}}}

\subsubsection{Generalization on Unseen Classes.} \quad
\label{sec:scalability}
We evaluate the proposed method on unseen classes to verify its generalization ability. We randomly choose some sketches from the Sketchy database~\cite{sketchy2016} (Fig.~\ref{fig:res_unseen}(Left)) and TU-Berlin sketch dataset~\cite{eitz2012hdhso} (Fig.~\ref{fig:res_unseen}(Right)) for testing.
We can see that although our model is trained on rigid object classes, it also performs well on non-rigid objects.
\section{Conclusion}
In this work, we have proposed a new method for sketch-based single-view 3D reconstruction. During sketch translation, an informative feature map is derived from an input sketch via an image translation model, which is then used to predict a density map for point cloud generation. The point cloud generation process is implemented by two-stage sampling strategy: with the guidance of the density map, the $x$ and $y$ coordinates are first recovered; and then based on the conditions of $x$ and $y$ coordinates and the input sketch the z coordinate is further predicted by sampling. Experimental results have demonstrated that our proposed method can significantly outperform the baseline methods.

\noindent\textbf{Acknowledgement:} \quad This work is supported by the National Natural Science Foundation of China (No. 62002012 and No. 62132001) and Key Research and Development Program of Guangdong Province, China (No. 2019B010154003).
\clearpage
%
%
\bibliographystyle{splncs04}
\bibliography{egbib}

\begin{thebibliography}{10}
\providecommand{\url}[1]{\texttt{#1}}
\providecommand{\urlprefix}{URL }
\providecommand{\doi}[1]{https://doi.org/#1}

\bibitem{bmvc/BianWLP21}
Bian, W., Wang, Z., Li, K., Prisacariu, V.A.: Ray-onet: Efficient 3d
  reconstruction from {A} single {RGB} image. In: British Machine Vision
  Conference(BMVC) (2021)

\bibitem{shapenet2015}
Chang, A.X., Funkhouser, T., Guibas, L., Hanrahan, P., Huang, Q., Li, Z.,
  Savarese, S., Savva, M., Song, S., Su, H., et~al.: Shapenet: An
  information-rich 3d model repository. arXiv preprint arXiv:1512.03012  (2015)

\bibitem{choy20163d}
Choy, C.B., Xu, D., Gwak, J., Chen, K., Savarese, S.: 3d-r2n2: A unified
  approach for single and multi-view 3d object reconstruction. In: ECCV (2016)

\bibitem{delanoy20183d}
Delanoy, J., Aubry, M., Isola, P., Efros, A.A., Bousseau, A.: 3d sketching
  using multi-view deep volumetric prediction. Proceedings of the ACM on
  Computer Graphics and Interactive Techniques(PACMCGIT)  \textbf{1}(1),  1--22
  (2018)

\bibitem{eitz2012hdhso}
Eitz, M., Hays, J., Alexa, M.: How do humans sketch objects? ACM TOG
  \textbf{31}(4),  44:1--44:10 (2012)

\bibitem{DBLP:journals/corr/FanSG16}
Fan, H., Su, H., Guibas, L.J.: A point set generation network for 3d object
  reconstruction from a single image. In: CVPR (2017)

\bibitem{gkioxari2019mesh}
Gkioxari, G., Malik, J., Johnson, J.: Mesh r-cnn. In: ICCV (2019)

\bibitem{goodfellow2014generative}
Goodfellow, I., Pouget-Abadie, J., Mirza, M., Xu, B., Warde-Farley, D., Ozair,
  S., Courville, A., Bengio, Y.: Generative adversarial nets. In: NeurIPS
  (2014)

\bibitem{guillard2021sketch2mesh}
Guillard, B., Remelli, E., Yvernay, P., Fua, P.: Sketch2mesh: Reconstructing
  and editing 3d shapes from sketches. In: ICCV (2021)

\bibitem{isola2017image}
Isola, P., Zhu, J.Y., Zhou, T., Efros, A.A.: Image-to-image translation with
  conditional adversarial networks. In: CVPR (2017)

\bibitem{kingma2014adam}
Kingma, D.P., Ba, J.: Adam: A method for stochastic optimization (2015)

\bibitem{li2017bendsketch}
Li, C., Pan, H., Liu, Y., Tong, X., Sheffer, A., Wang, W.: Bendsketch: modeling
  freeform surfaces through 2d sketching. ACM TOG  \textbf{36}(4),  1--14
  (2017)

\bibitem{li2018robust}
Li, C., Pan, H., Liu, Y., Tong, X., Sheffer, A., Wang, W.: Robust flow-guided
  neural prediction for sketch-based freeform surface modeling. ACM TOG
  \textbf{37}(6),  1--12 (2018)

\bibitem{lin2017feature}
Lin, T.Y., Doll{\'a}r, P., Girshick, R., He, K., Hariharan, B., Belongie, S.:
  Feature pyramid networks for object detection. In: CVPR (2017)

\bibitem{lun20173d}
Lun, Z., Gadelha, M., Kalogerakis, E., Maji, S., Wang, R.: 3d shape
  reconstruction from sketches via multi-view convolutional networks. In: 3DV
  (2017)

\bibitem{mescheder2019occupancy}
Mescheder, L., Oechsle, M., Niemeyer, M., Nowozin, S., Geiger, A.: Occupancy
  networks: Learning 3d reconstruction in function space. In: CVPR (2019)

\bibitem{mirza2014conditional}
Mirza, M., Osindero, S.: Conditional generative adversarial nets. arXiv
  preprint arXiv:1411.1784  (2014)

\bibitem{nguyen2019graphx}
Nguyen, A.D., Choi, S., Kim, W., Lee, S.: Graphx-convolution for point cloud
  deformation in 2d-to-3d conversion. In: ICCV (2019)

\bibitem{eccv/PopovBF20}
Popov, S., Bauszat, P., Ferrari, V.: Corenet: Coherent 3d scene reconstruction
  from a single {RGB} image. In: ECCV (2020)

\bibitem{sketchy2016}
Sangkloy, P., Burnell, N., Ham, C., Hays, J.: The sketchy database: learning to
  retrieve badly drawn bunnies. ACM TOG  \textbf{35}(4),  1--12 (2016)

\bibitem{shin2018pixels}
Shin, D., Fowlkes, C.C., Hoiem, D.: Pixels, voxels, and views: A study of shape
  representations for single view 3d object shape prediction. In: CVPR (2018)

\bibitem{wang20203d}
Wang, J., Lin, J., Yu, Q., Liu, R., Chen, Y., Yu, S.X.: 3d shape reconstruction
  from free-hand sketches. arXiv preprint arXiv:2006.09694  (2020)

\bibitem{wang2018unsupervised}
Wang, L., Qian, C., Wang, J., Fang, Y.: Unsupervised learning of 3d model
  reconstruction from hand-drawn sketches. In: ACM MM (2018)

\bibitem{wang20173densinet}
Wang, M., Wang, L., Fang, Y.: 3densinet: A robust neural network architecture
  towards 3d volumetric object prediction from 2d image. In: ACM MM (2017)

\bibitem{wang2018pixel2mesh}
Wang, N., Zhang, Y., Li, Z., Fu, Y., Liu, W., Jiang, Y.G.: Pixel2mesh:
  Generating 3d mesh models from single rgb images. In: ECCV (2018)

\bibitem{wu2017marrnet}
Wu, J., Wang, Y., Xue, T., Sun, X., Freeman, B., Tenenbaum, J.: Marrnet: 3d
  shape reconstruction via 2.5d sketches. In: NeurIPS (2017)

\bibitem{xie2019pix2vox}
Xie, H., Yao, H., Sun, X., Zhou, S., Zhang, S.: Pix2vox: Context-aware 3d
  reconstruction from single and multi-view images. In: ICCV (2019)

\bibitem{ijcv/XieYZZS20}
Xie, H., Yao, H., Zhang, S., Zhou, S., Sun, W.: Pix2vox++: Multi-scale
  context-aware 3d object reconstruction from single and multiple images. IJCV
  \textbf{128}(12),  2919--2935 (2020)

\bibitem{Xu:2014:True2Form}
Xu, B., Chang, W., Sheffer, A., Bousseau, A., McCrae, J., Singh, K.: True2form:
  3d curve networks from 2d sketches via selective regularization. ACM TOG
  \textbf{33}(4) (2014)

\bibitem{xu2019disn}
Xu, Q., Wang, W., Ceylan, D., Mech, R., Neumann, U.: {DISN:} deep implicit
  surface network for high-quality single-view 3d reconstruction. In: NeurIPS
  (2019)

\bibitem{zhang2021sketch2model}
Zhang, S.H., Guo, Y.C., Gu, Q.W.: Sketch2model: View-aware 3d modeling from
  single free-hand sketches. In: CVPR (2021)

\bibitem{zhong2020deep}
Zhong, Y., Gryaditskaya, Y., Zhang, H., Song, Y.Z.: Deep sketch-based modeling:
  Tips and tricks. In: 3DV (2020)

\bibitem{zhong2020towards}
Zhong, Y., Qi, Y., Gryaditskaya, Y., Zhang, H., Song, Y.Z.: Towards practical
  sketch-based 3d shape generation: The role of professional sketches. IEEE
  Transactions on Circuits and Systems for Video Technology(T-CSVT)
  \textbf{31}(9),  3518--3528 (2020)

\end{thebibliography}
\end{document}